\pdfoutput=1

\documentclass[11pt]{article}

\usepackage[final]{coling}

\usepackage{times}
\usepackage{latexsym}

\usepackage[T1]{fontenc}

\usepackage[utf8]{inputenc}

\usepackage{microtype}

\usepackage{inconsolata}

\usepackage{graphicx}
\usepackage{amsmath,amsthm,amssymb,bm}
\usepackage{booktabs}
\usepackage{multirow,makecell,multicol}
\newcommand{\forward}[1]{\overrightarrow{#1}}
\newcommand{\backward}[1]{\overleftarrow{#1}}
\usepackage[inline]{enumitem}

\usepackage{wrapfig}

\title{Acquiring Bidirectionality via Large and Small Language Models}

\author{
  \textbf{Takumi Goto\textsuperscript{1,2}},
  \textbf{Hiroyoshi Nagao\textsuperscript{1}},
  \textbf{Yuta Koreeda\textsuperscript{1}},
\\
  \textsuperscript{1}Research \& Development Group, Hitachi, Ltd., Tokyo, Japan,
\\
  \textsuperscript{2}NARA Institute of Science and Technology, Nara, Japan,
\\
  \small{
    \textbf{Correspondence:} \href{mailto:yuta.koreeda.pb@hitachi.com}{yuta.koreeda.pb@hitachi.com}
  }
}

\begin{document}
\maketitle
\begin{abstract}
Using token representation from bidirectional language models (LMs) such as BERT is still a widely used approach for token-classification tasks.
Even though there exist much larger unidirectional LMs such as Llama-2, they are rarely used to replace the token representation of bidirectional LMs.
In this work, we hypothesize that their lack of bidirectionality is what is keeping unidirectional LMs behind. 
To that end, we propose to newly train a small backward LM and concatenate its representations to those of an existing LM  for downstream tasks.
Through experiments in token-classification tasks, we demonstrate that introducing a backward model can improve the benchmark performance by more than 10 points.
Furthermore, we show that the proposed method is especially effective for rare domains and in few-shot learning settings.
\end{abstract}

\section{Introduction}
In recent years, pretrained large unidirectional language models (UniLMs), such as Llama-2~\cite{touvron2023llama} and OpenAI GPT~\cite{openai2024gpt4}, have become widely used.
Large UniLMs have demonstrated that various tasks can be solved by means of generation.
On the other hand, bidirectional language models (BiLMs), most well-known by BERT~\cite{devlin-etal-2019-bert}, equipped with a classification layer are still widely used in many NLP tasks.
In particular, BiLMs are still dominant for token-level classification tasks.
For example, as of 2024, top three models in two popular token-level classification tasks, CoNLL 2003 named entity recognition (NER)~\cite{tjong-kim-sang-de-meulder-2003-introduction} and DocRED relationship extraction~\cite{yao-etal-2019-docred}, are all based on BiLMs\footnote{From Papers With Code, as of May, 2024. Top three models are \cite{wang-etal-2021-automated,yamada-etal-2020-luke,zhou-chen-2021-learning} for \href{https://paperswithcode.com/sota/named-entity-recognition-ner-on-conll-2003}{CoNLL 2003} and \cite{ma-etal-2023-dreeam,tan-etal-2022-document,Xu_Wang_Lyu_Zhu_Mao_2021} for \href{https://paperswithcode.com/sota/relation-extraction-on-docred}{DocRED}.}.

The reason why the application of large UniLMs in token-level classification tasks has not progressed can be attributed to their lack of bidirectionality.
In a UniLM, the representation of a token is computed solely based on the preceding context, as we elaborate in Section~\ref{sec:preliminary}.
To overcome this problem, \citet{behnamghader2024llm2vec} introduced LLM2Vec, where UniLMs are fine-tuned with masked token prediction after removing their causal attention masks.
This allows the model to attend to both the beginning and the end of a sentence thus acquiring bidirectionality.
This allows utilizing existing UniLMs not only for generation tasks but also for highly accurate solutions to token-level classification tasks.

LLM2Vec, however, has a downside that it requires training for each UniLM, which is costly if we are to try out various UniLMs to find a good fit for a downstream task.
Given the already large and rapidly evolving zoo of UniLMs, it would be beneficial if there is a one-for-many solution for equipping bidirectionality to UniLMs.

In this work, we propose a new way to acquire bidirectionality without tuning large UniLMs themselves.
Specifically, we newly train a small UniLM for generating text from the end (referred to the ``backward LM'') and concatenate its token representations to those of the pretrained UniLM (referred to the ``forward LM'') to obtain pseudo bidirectionality.
After that, we train only the classification layer for the downstream tasks as a drop-in replacement of BiLMs.
The backward LM is independent on which forward LM it is used with, thus it can be combined with various size of UniLMs even if it ended up in a heterogeneous configuration.

In the experiments, we focus on three kinds of token-classification tasks, i.e., chunking, part-of-speech (POS) tagging and NER, and compare the performances of UniLMs with and without backward LM.
We observe that adding backward LMs consistently improves the performance by up to more than 10 points in CoNLL2003-NER.
Additionally, we demonstrate that the proposed method consistently improves performance in few-shot settings or when targeting rare domains.

The contributions of this study are as follows:
\begin{enumerate}
    \item We empirically show that unidirectionality is a problem when adopting UniLMs to token-level classification tasks.
    \item We proposed a novel method to newly train a small-scale backward LM and concatenate its representations to those of existing LM to achieve pseudo bidirectionality in UniLMs.
    \item We open-sourced backward LM and its training code to foster future research\footnote{\url{https://github.com/hitachi-nlp/backward-llm}}.
\end{enumerate}

\section{Proposed Method}
\subsection{Prerequisite}\label{sec:preliminary}

In this section, we review two types of LMs: UniLMs and BiLMs.
Given an input sequence $\boldsymbol{x} = (x_1, x_2, \dots, x_N)$ with $N$ tokens, the difference between the two models lies in how they compute representations for each word $x_i$ $(1 \leq i \leq N)$.

In BiLMs, the representation $\mathbf{h}_i^{bi}$ is computed based on the context from both the beginning and the end of the text:
\begin{equation}\label{eq:bidirectional}
    \mathbf{h}_i^{bi} = BiLM_{\phi}(x_i | \boldsymbol{x}_{<i}, \boldsymbol{x}_{>i}),
\end{equation}
where $\boldsymbol{x}_{<i} = (x_1, x_2, \dots, x_{i-1})$ and $\boldsymbol{x}_{>i} = (x_{i+1}, x_{i+2}, \dots, x_N)$.
To solve token-classification tasks, we can input $\mathbf{h}_i^{bi}$ to the newly added classification layers.

On the other hand, a forward LM computes the representation $\forward{\mathbf{h}}_i$ solely based on earlier context:
    \begin{equation}\label{eq:unilm-forward}
        \forward{\mathbf{h}}_i = \forward{Uni}LM_{\theta}(x_i | \boldsymbol{x}_{<i}).
    \end{equation}
As can be seen from equation \eqref{eq:unilm-forward}, UniLMs need to compute the representation for the $i$-th word without using the subsequent context $\boldsymbol{x}_{>i}$.

\subsection{Utilization of the Bidirectional Language Model}

The proposed method leverages the concatenated representations of both the forward LM $\forward{Uni}LM(\cdot)$ and the backward LM $\backward{Uni}LM(\cdot)$ for the downstream task.
In contrast to the forward LM, a backward LM computes $\backward{\mathbf{h}_i}$ given the context from the end:

\begin{equation}\label{eq:unilm-backward}
    \backward{\mathbf{h}}_i = \backward{Uni}LM_{\theta'}(x_i | \boldsymbol{x}_{>i}).
\end{equation}

The final representation for the $i$-th token considers both the forward and backward contexts by concatenating $\forward{\mathbf{h}_i}$ and $\backward{\mathbf{h}_i}$, denoted as $\mathbf{h}_i = \mathrm{Concat}[\forward{\mathbf{h}_i}, \backward{\mathbf{h}_i}]$. 
Therefore, the dimensions of $\mathbf{h}_i$ is the sum of the hidden vector dimensions of the forward and backward LMs.

To compute the concatenated representations as described above, it is necessary to share the vocabulary between the forward and backward LMs.
Nevertheless, it is possible to use arbitrary architectures and parameter sizes for both models.
For instance, we can employ a heterogeneous configuration, such as $|\theta| \gg |\theta'|$.
This means that we can utilize the existing assets of $|\theta|$ with just a small compute of training $|\theta'|$.
In this study, a part of experiments is conducted with such heterogeneous configuration.

\section{Experiments}

We verify whether UniLMs can acquire bidirectionality by adding backward UniLMs through evaluation on four token-classification tasks.
We train a backward LM (124M parameters) and apply it to GPT-2 (base, 124M)~\cite{radford2019language} and Llama2-7b~\cite{touvron2023llama}, to verify whether the proposed method can be applied to UniLMs of different sizes.

\subsection{Training Backward LM}

\begin{figure*}[t]
  \centering
  \includegraphics[width=1\linewidth]{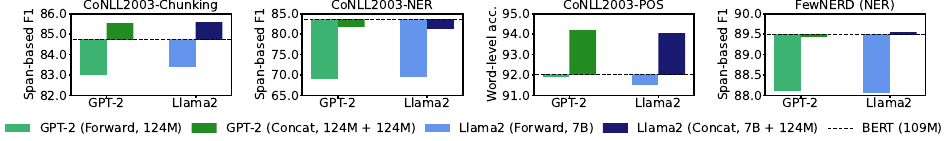}
  \caption{The performance of UniLMs (``Forward'') and the proposed concatenated LMs (``Concat'') against a BiLM (BERT)}
  \label{fig:results}
\end{figure*}

We train a backward LM for each of Llama2 and GPT-2, as the backward LM should have the same vocabulary as the forward LM.
The architecture follows that of GPT-2 (base), but we resize the input dimension of the embedding layer to match the vocabulary size of the forward LM.
We initialize the models with random parameters and train it on BookCorpus~\cite{Zhu_2015_ICCV} and Wikitext~\cite{merity2016pointer} (\texttt{wikitext-103-raw-v1}) datasets, with next token prediction objective.
During the preprocessing step, we concatenate the training data from both datasets and shuffle them on a document level\footnote{For Wikitext, we removed empty lines and strings corresponding to headings beforehand.}.
Next, we perform subword tokenization with the forward LM's tokenizer.
We extract training data by cutting it into segments of 1,024 tokens, starting the beginning of the dataset, and then reversing them.
For training, we set the batch size to 512 and the learning rate to 2e-5 with a cosine scheduler.

\subsection{Downstream Tasks}\label{sec:ner}

To evaluate the effectiveness of the proposed method in a token-level classification task, we employ chunking, POS tagging, NER from CoNLL-2003 dataset~\cite{tjong-kim-sang-de-meulder-2003-introduction}.
In addition, we utilize Few-NERD dataset (supervised setting)~\cite{ding-etal-2021-nerd} to verify the NER performance on rare domains.
We hypothesize that larger forward LMs are effective in this more challenging setting because of their extensive knowledge.

In the evaluation, we use span-based $F_1$ score for chunking and NER, and word-level accuracy for POS. 

During downstream task training, we input the representations $\mathbf{h}_i$ from each model into the classification layer and optimize the classification layer while keeping using cross-entropy loss.
The classification layer consists of two linear layers.
Let $d$ be the dimensionality of $\mathbf{h}_i$ and $c$ be the number of classes.
We use $\mathbf{W}_{1} \in \mathbb{R}^{d \times d}$ and $\mathbf{W}_{2} \in \mathbb{R}^{c \times d}$ to estimate the distribution $\mathbf{p} = \mathrm{softmax}(\mathbf{W}_{2}\mathrm{ReLU}(\mathbf{W}_{1}\mathbf{h}_i)) \in \mathbb{R}^{c}$.

We set the batch size to 32 and employ AdamW~\cite{loshchilov2019decoupled}.
We linearly decay the learning rate to zero from 1e-3.
We only train the classification layer while keeping the other layers fixed.

We report the average scores of three different seeds using the checkpoint that has maximum $F_{1}$ scores on the validation set.
We also train a model using BERT (\texttt{bert-base-uncased}) with the same setting to compare the proposed method to a BiLM.

\subsection{Few-shot Setting}

One of the potential benefits of large LMs is their ability to make generalized predictions even with a small number of training examples, leveraging the knowledge embedded in their parameters.
We also analyze NER performance on CoNLL-2003 with $K$-shot setting to examine the impact of limited training examples.

In our $K$-shot setting, the training data consists of $4K$ samples since we draw $K$ samples from each entity type: \texttt{PER}, \texttt{LOC}, \texttt{ORG}, and \texttt{MISC}.
Note that we only extract instances from the training data that contain a single specific named entity type.
During training, we set the batch size to 4 and randomly sampled the following hyperparameters; a learning rate from $\{9e-3, 8e-3, \dots, 2e-4, 1e-4\}$ , a seed from $\{10, 11, \dots, 19\}$, a dropout probability from $\{0, 0.1, 0.2, 0.3\}$.
We determined top-3 hyperparameter settings in terms of $F_1$ score on the CoNLL-2003 validation set, and report the average $F_1$ on the test set of those settings.

\subsection{Results}\label{sec:results}

\subsubsection{Full Dataset Setting}
We show the results in Figure~\ref{fig:results}.
The results show the difference of the performance between BERT and each of the settings.
We can see the effectiveness of the proposed method by comparing ``Forward'' and ``Concat'' settings, which indicates the proposed method improves the performance on all tasks.
In particular, $F_1$ scores on CoNLL2003-NER have been improved by more than 10 points for both models.

These results indicate that considering backward context improves token-classification performance and that the proposed method can provide the backward context to UniLMs.
Moreover, the comparison of BERT and the proposed method indicates that our approach has ability to bring the performance of existing UniLMs comparable or better to BERT performance.
This implies that UniLMs can acquire bidirectionality post hoc.

In the case of Few-NERD, Llama-2 outperformed GPT-2 with the proposed method.
It can be inferred that larger forward LMs are more effective when targeting a rarer domain.

\begin{figure}[t]
  \centering
  \includegraphics[width=0.95\linewidth]{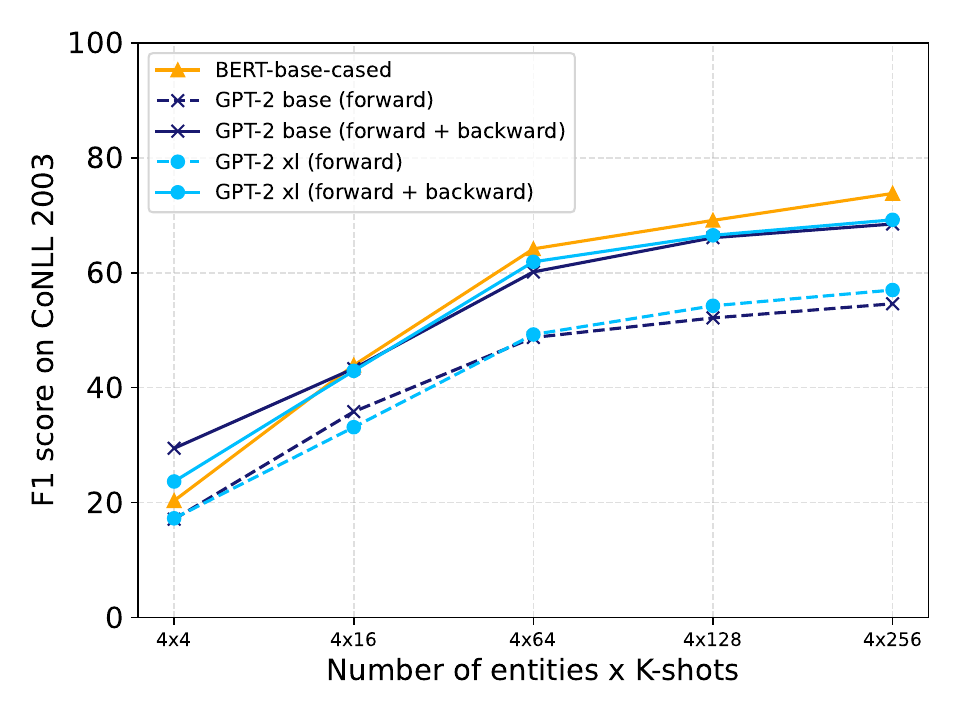}
  \caption{Few-shot setting results on CoNLL-2003 test set. In the $x$-axis, the number of training examples is represented by multiplication of the number of entities ($=4$) and $K$.}
  \label{fig:few-conll2003}
\end{figure}

\subsubsection{Few-shot Setting}

The experimental results using BERT (\texttt{bert-base-cased}) and GPT-2 (base and xl) are shown in Figure~\ref{fig:few-conll2003}. 
The $x$-axis represents the number of training examples $4K$, and the $y$-axis represents the $F_1$ score on CoNLL2003 test set.
The dashed lines indicate only forward LM while the solid lines represent the bidirectional and the proposal setting.
As observed from Figure~\ref{fig:few-conll2003}, the proposed method consistently outperforms the forward LM-only setting. 
Particularly, when the training data is limited (less than 16 shots for each entity) the proposed method is more effective than BERT.
This has a significant value in practice as there is generally a \emph{valley} of performance between few- to many-shots settings;
zero- to few-shots settings are more effectively addressed by in-context learning and many-shots settings are covered by BiLMs, but neither covers the middle.
Addressing this \emph{valley} is important, because annotating dozens of data for each label might be justified but annotating hundreds sounds overwhelming to many practitioners.

\subsection{Case Study}
We conduct a case study to understand how adding backward context improves the token-level classification performance.
As shown in Table~\ref{tab:example1}, we found that the proposed method is particularly effective when there is an entity at the beginning of a sentence.
Specifically, the NER results for the sentence ``Jones Medical completes acquisition .'' are [B-PER, I-ORG, --, --, --] for the forward UniLM setting only, and [B-ORG, I-ORG, --, --, --] for the proposed method and the reference.
The forward UniLM could not capture any context because the entity appears at the beginning of the sentence.
In contrast, the proposed method was able to predict the entity using the context from the end.
We also found that the proposed method could accurately estimate the leading entity in phrases where entities are conjoined by ``and.''\footnote{The actual example can be found in Appendix~\ref{sec:appendix-example}.}
These results suggest that the UniLM representations are not suitable for token-level classification tasks, even when the LM is of large size, but the proposed method is able to overcome this weakness by the simple idea.

\begin{table}[t]
\centering
\fontsize{8pt}{10pt}\selectfont
\renewcommand{\arraystretch}{.8}
\setlength{\tabcolsep}{5.5pt}
\begin{tabular}{l|ccccc}
	\toprule
	Input &  Jones & Medical & completes & acquisition & . \\
	\midrule
	UniLM & B-PER & I-ORG & - & - & - \\
	Proposal & B-ORG & I-ORG & - & - & - \\
	Reference & B-ORG & I-ORG & - & - & - \\
	\bottomrule
\end{tabular}
 \caption{An example of NER with GPT-2 (base) in a case that an entity at the beginning of the sentence.}
 \label{tab:example1}
\end{table}

\section{Related Work}
A traditional approach to combining UniLMs, such as BiLSTM~\cite{schuster1997bidirectional} and ELMo~\cite{peters-etal-2018-deep}, is similar to the proposal method.
Our study revisits this idea in the era of large LMs, and systematically shows its effectiveness through compute-demanding experiments.
Our method is simple, but it in returns shows that backward context matters even in the era of large LM and that the traditional approach of acquiring bidirectionality is still valid.
In the era of large language models, \emph{meet-in-the-middle} approach \cite{nguyen2023meet,li2023batgpt} consider bidirectionality during generation by incorporating backward generation probability.
While there is some relevancy in the concept, these studies work with generation whereas we aim to improve the quality of token-level representations.

For the study to improve the quality of the representation, \citet{li2023label}, \citet{behnamghader2024llm2vec} and \citet{dukic-snajder-2024-looking} fine-tuned UniLMs after removing the causal attention mask to incorporate the context from the end of the sentence.
Although these methods require fine-tuning for each LM, the proposed method can reuse the backward LM as long as the vocabulary and tokenizer are the same.
Moreover, another benefit is that our approach can be applied to a black-box model, i.e., when the parameters of a model are not accessible but its final representations can be obtained via API.

\section{Conclusion}
In this study, we proposed to concatenate the representations of forward and backward LM, to overcome the lack of bidirectionality problem of UniLMs.
From the results in token-classification tasks, we could confirm the effectiveness of the proposed method.
The proposed method provides more use cases to UniLM, not only for a generation model but also as an encoder model.

\section*{Acknowledgments}

We would like to thank Dr. Masaaki Shimizu for arranging the computational environment.
We would like to thank the COLING reviewers and program chairs for their suggestive comments and feedbacks.

\section*{Limitations}

\paragraph{Limited Scope of Tasks and Conditions}

In this work, we evaluated our proposed method with two backbone UniLMs under four different token-level classification tasks in English, and we observed consistent performance improvement from the original UniLMs.
We focused on token-level classification tasks in this study because we believe they are the tasks where the effects of considering bidirectionality are best demonstrated.
Nevertheless, we would like to extend the experiments in the future to investigate varying effects of the proposed method under different conditions.
For example, we can apply the proposed method to text classification tasks by pooling token-level representations.
Also, we would like to see if significantly larger UniLMs can yield better results, as we did not observe any performance gain within the relatively small scope of model sizes (124M to 7B parameters) that we experimented with in this paper.

\paragraph{Training Strictly Comparable Models from Scratch}

We compared BERT and GPT-2 because\begin{enumerate*}[label=(\alph*)]
  \item they were release roughly the same time and together represent the state-of-the-arts of unidirectional and bidirectional LM of the time, and
  \item they are mostly comparable in terms of magnitude of parameter sizes and training data
\end{enumerate*}.
In reality, there \emph{might} be auxiliary differences because the dataset for training or detail training settings, e.g., the number of epochs, to train GPT-2 are not disclosed~\cite{radford2019language}.
Though our main finding that unidirectionality is a problem in UniLMs holds from the GPT-2 results alone, strict comparison can be performed by training BERT and GPT-2-like models from scratch with the same trainig data, which we leave for the future work.

Additionally, we can also consider using BiLMs instead of a backward language model, as a provider of backward context.

\paragraph{Comparison against Other Methods}

We did not quantitatively compare the proposed method against LLM2vec~\cite{behnamghader2024llm2vec} in the experiments.
It was due to the fact their reported results employed non-standard word-level scores for NER instead of more standard span-level scores that we employed and hence direct comparison was not possible.
We note that our method has unique benefits that LLM2vec does not have: \begin{enumerate*}[label=(\alph*)]
    \item our backward model can be reused for different backbone UniLMs as long as the tokenizer is the same, and
    \item we only need the output representations of UniLMs and do not require access to the internal representations nor gradients of the UniLMs 
\end{enumerate*}.
Furthermore, our work has contribution to the community by validating through a different approach that bidirectionality is the key ingredient of the success of BiLMs.

\paragraph{Computation Cost}

The proposed method requires a backward LM, thus the inference computation cost is higher than when using only a forward LM.
Nevertheless, this is generally not a serious problem because the backward language model works even with modest sized models such as GPT-2.
In practice, the proposed method can be executed by reusing the large UniLMs deployed for other purpose, e.g., generation, so there should be cases where it is sufficient to deploy only the backward UniLM.
In this case, unless the UniLM's GPU utilization is 100\%, we can run the proposed method with only the cost of running backward UniLM.

\section*{Ethical Consideration}

Our research involves training moderately large LMs and its carbon footprint can have negative impact to the environment.
That being said, we train reusable LMs that allow us to utilize existing assets (large UniLMs) to where they were previously weak at.
This can make the whole ecosystem of LLMs more efficient and might be able to reduce carbon footprint in the long run.

From the social justice and language preservation perspective, the downside of the proposed method is that it mainly benefits resource-rich languages with existing assets of large UniLMs.
Nevertheless, more and more of recent UniLMs support multilinguality~\cite{gemmateam2024gemma2improvingopen,dubey2024llama3herdmodels}.
We would like to explore our proposed method in low-resource, cross-lingual settings in the future.

\begin{table*}[t]
	\centering
	\small
	\begin{tabular}{l|ccccccccccccc}
		\toprule
		Input &  Note & - & Lotte & and & Hyundai & , & Haitai & and & Samsung & played & two & games & . \\
		\midrule
		Forward-LM only & - & - & B-PER & - & B-ORG & - & - & & B-ORG & - & - & - & - \\
		Proposed & - & - & B-ORG & - & B-ORG & - & B-MISC & & B-ORG & - & - & - & -  \\
		Reference & - & - & B-ORG & - & B-ORG & - & B-ORG & & B-ORG & - & - & - & -  \\
		\bottomrule
	\end{tabular}
	\caption{An example of GPT-2 (base) in phrases where entities are conjoined by ``and.'' For the prediction corresponding to \texttt{Lotte}, the forward-LM struggles to infer the entity type correctly, but proposal can estimate correctly. This can be explained by the difference between considering only the context from the beginning: ``\texttt{Note - Lotte}'' or entire context, cause the improvement.}
	\label{tab:example2}
\end{table*}

\bibliography{custom}

\newpage

\section*{Appendices}
\appendix

\section{Another Example in Case Study}\label{sec:appendix-example}

Table~\ref{tab:example2} shows an example where entities are concatenated with ``and.''
The forward-LM only setting struggles with identifying the type of first entity, for ``Lotte.''
In contrast, the proposed method correctly identified the entity type of it by using other organization names such as ``Hyundai.''

\end{document}